\documentclass[conference, table]{IEEEtran}
\IEEEoverridecommandlockouts
\usepackage{cite}
\usepackage{amsmath,amssymb,amsfonts}
\usepackage{algorithmic}
\usepackage{balance}
\usepackage{flushend}
\usepackage{graphicx}
\usepackage{textcomp}
\usepackage{xcolor}
\def\BibTeX{{\rm B\kern-.05em{\sc i\kern-.025em b}\kern-.08em
    T\kern-.1667em\lower.7ex\hbox{E}\kern-.125emX}}
\usepackage{tikz}
\usepackage{hyperref}
\usepackage{tcolorbox}
\definecolor{fhgreen}{RGB}{0,147,116}
\definecolor{orange}{RGB}{	255, 127, 0}
\usepackage{mathtools}
\usepackage{amsthm}
\usepackage{bbm}
\usepackage{bm}
\usepackage{tabularx}
\usepackage{threeparttable}
\usepackage{makecell}
\usepackage{multirow}
\usepackage{booktabs}
\usepackage{subcaption}
\usepackage{flushend}

\newcommand{\RR} {\mathbb R}

\newcommand{\mat}{\bm}

\makeatletter
\def\ps@IEEEtitlepagestyle{%
  \def\@oddfoot{\mycopyrightnotice}%
}
\def\mycopyrightnotice{%
\fbox{\parbox{\dimexpr\textwidth-2\fboxsep-2\fboxrule\relax}{
\begin{minipage}{\textwidth-2\fboxsep-2\fboxrule}
  \footnotesize
  \textcopyright 2022 IEEE. Personal use of this material is permitted. Permission from IEEE must be obtained for all other uses, in any current or future media, including reprinting/republishing this material for advertising or promotional purposes, creating new collective works, for resale or redistribution to servers or lists, or reuse of any copyrighted component of this work in other works.
  \end{minipage}
}}
}
\makeatother

\begin{document}

\title{
KPI-EDGAR: A Novel Dataset and Accompanying Metric for Relation Extraction from Financial Documents
\thanks{This research will be published in the 2022 proceedings of the IEEE International Conference on Machine Learning and Applications (ICMLA) and has been funded by the Federal Ministry of Education and Research of Germany and the state of North-Rhine Westphalia as part of the Lamarr-Institute for Machine Learning and Artificial Intelligence, LAMARR22B.}
}

\author{
    \IEEEauthorblockN{
        Tobias Deußer\IEEEauthorrefmark{1}\IEEEauthorrefmark{2}, Syed Musharraf Ali\IEEEauthorrefmark{2}\IEEEauthorrefmark{3}, Lars Hillebrand\IEEEauthorrefmark{1}\IEEEauthorrefmark{2}, Desiana Nurchalifah\IEEEauthorrefmark{2}\IEEEauthorrefmark{3}, Basil Jacob\IEEEauthorrefmark{2}, \\Christian Bauckhage\IEEEauthorrefmark{1}\IEEEauthorrefmark{2}, Rafet Sifa\IEEEauthorrefmark{2}
    }
    \IEEEauthorblockA{\IEEEauthorrefmark{1} University of Bonn, Bonn, Germany}
    \IEEEauthorblockA{\IEEEauthorrefmark{2} Fraunhofer IAIS, Sankt Augustin, Germany}
    \IEEEauthorblockA{\IEEEauthorrefmark{3} Hochschule Bonn-Rhein-Sieg, Sankt Augustin, Germany}
    \texttt{tobias.deusser@iais.fraunhofer.de}
    \\ \texttt{ORCID iD: 0000-0003-4685-0847}
}

\maketitle

\begin{abstract}
We introduce KPI-EDGAR, a novel dataset for Joint Named Entity Recognition and Relation Extraction building on financial reports uploaded to the Electronic Data Gathering, Analysis, and Retrieval (EDGAR) system, where the main objective is to extract Key Performance Indicators (KPIs) from financial documents and link them to their numerical values and other attributes. We further provide four accompanying baselines for benchmarking potential future research. Additionally, we propose a new way of measuring the success of said extraction process by incorporating a word-level weighting scheme into the conventional F$_1$ score to better model the inherently fuzzy borders of the entity pairs of a relation in this domain.
\end{abstract}


\section{Introduction}
Financial documents like income statements, business plans, initial public offering prospectuses, or whole annual financial reports hold a large amount of information for the reader willing to spend a considerable time reading through and analyzing them. Financial auditors are employed to check their validity and financial analysts utilize them to evaluate companies. Thus, creating an algorithm that automatically extracts information is likely to give the user a competitive edge by saving them extensive time usually spend on in-depth reading. 

Arguably, one of the most important pieces of information contained in such documents are key performance indicators (KPIs), which are defined as ``vital navigation instruments used by managers to
understand whether their business is on a successful voyage or whether it is veering off the
prosperous path'' \cite{marr2012key}, and are mostly quantitative measures. Well-known examples are revenue, profit, and loss.

In a previous paper, \cite{kpi-bert} introduced KPI-BERT, a system which employed novel methods of named entity recognition (NER) and relation extraction (RE) to extract and link key performance indicators of companies from real-world German financial documents. This model was trained on undisclosed annotations in public financial documents owned by a major auditing company, and thus, unavailable to the research community.

To alleviate this flaw and to facilitate reproducibility as well as further research, we introduce KPI-EDGAR, a dataset based on publicly available data and manually annotated by us, with the main difference being that the language changed from German to English. It was collected from the Electronic Data Gathering, Analysis, and Retrieval system\footnote{At the time of writing, the EDGAR database can be accessed at this link:
\href{https://www.sec.gov/edgar.shtml}{sec.gov/edgar}}, usually shortened to EDGAR, which is maintained by the U.S. Securities and Exchange Commission and is made up of annual comprehensive financial reports, so-called \textit{10-K} reports.  It will be published with the release of this paper.

Furthermore, we present an adjusted F$_1$ metric, that considers an entity and thus relation not as a unified, single object, but as a collection of words that should be evaluated as such. This is based on the observation that a relation extraction algorithm is often partly correct and missing only a minor part of the whole relation, regularly a rather unimportant word. The conventional F$_1$ metric would still assign zero accuracy to this prediction, hence not reflecting the actual performance. These prediction errors probably stem from the fact that the word boundaries of a particular KPI might not be clearly defined and even financial auditors can be unsure where the exact confines of a KPI lie. To mitigate this issue, we incorporate a weighting scheme into the F$_1$ metric to better capture these cases.

To sum up, our contributions are:
\begin{itemize}
    \item We introduce a novel dataset, named KPI-EDGAR, for joint named entity recognition and relation extraction in the financial text domain based on actual corporate reports submitted to the EDGAR database. In contrast to other resources scraped from EDGAR (see \cite{han2016mining}, \cite{ashraf2017scraping}, \cite{loukas2021edgar}, and \cite{lonare2021edgar}), we provide manual annotations along with the corpus to allow for joint named entity recognition and relation extraction.
    \item We propose a new metric for measuring the performance of the relation extraction task to better capture when predictions are partially correct.
\end{itemize}

The rest of this paper is structured as follows. We first review related work. In the next section, we describe our novel dataset, KPI-EDGAR, in detail and how it was gathered and pre-processed. 
Section \ref{section:methodology} defines the adjusted metric that incorporates a weighting scheme and introduces and specifies the four baselines we provide to researchers to benchmark their approaches. Thereafter, we evaluate these four baselines on KPI-EDGAR and report the conventional and adjusted F$_1$ metric as well as a few example predictions to highlight strengths and weaknesses of the metric. We close this paper with concluding remarks and take an outlook into conceivable future work.

\section{Related Work}
Similar to our topic, many recent studies have focused on named entity recognition (\cite{lester2020constrained}, \cite{ushio2021t}, \cite{wang2021improving},  \cite{wang2021automated}), relation extraction (~\cite{xu2021entity}, \cite{zeng2020double}, \cite{zhangdocument}), and the joint combination of both (\cite{li2014incremental}, \cite{pappu2017lightweight}, \cite{eberts2019span}, \cite{liang2020bond}, \cite{shen2021trigger}, \cite{wang2020two}, \cite{ye2021pack}, \cite{zhong2021frustratingly}, \cite{tran2021improved}, \cite{hang2022joint}, \cite{savary2022relation}). Furthermore, \cite{fundel2007relex} as well as \cite{gurulingappa2012extraction} have shown that hierarchically using both tasks in a single pipeline can significantly improve the performance thus indicating that information from one task can be exploited by the other.
Additionally, methods described in recent publications are usually based on modern pre-trained natural language models (\cite{devlin2018bert}, \cite{liu2019roberta}, \cite{radford2018improving}, \cite{radford2019language}) and these methods have produced state-of-the-art results as expected.

For extracting information from financial reports, \cite{treigueiros1991application} provided pre-available accounting variables to a multilayer perceptron as input to build interpretable structures similar to accounting ratios. In comparison, \cite{farmakiotou2000rule} automatically extracted these variables and developed a NER model using a rule-based approach. \cite{brito2019hybrid} developed a hybrid AI tool that extracts KPIs by first detecting paragraphs and tables with computer vision techniques and then leverage supervised decision tree-based classifiers to accomplish the final extraction process.
 \cite{kpi-bert} applied a pre-trained language model combined with conditional label masking to extract key performance indicators from German financial reports. \cite{cao2018towards} used a joint entity and relation extraction approach to cross-check different financial formulas in Chinese documents.

Regarding the herein proposed metric, \cite{segura2013semeval} proposed a similar setup, but for Named Entity Recognition. Furthermore, \cite{fu2020interpretable} suggested dividing entities into buckets and evaluating the performance on each of these buckets.

Due to EDGAR's popularity, many researchers (\cite{han2016mining}, \cite{ashraf2017scraping}, \cite{lonare2021edgar}) have developed methods to extract data from EDGAR and used it in their own research. \cite{loukas2021edgar} even released a comprehensive corpus comprising annual reports from all the publicly traded companies in the US spanning a period of more than 25 years and accompanied it with a word2vec \cite{word2vec} model titled EDGAR--W2V, which we will also use as a baseline in our experiments. Furthermore, \cite{mustafaoglu2022lstm}, \cite{ryans2021textual}, and \cite{li2010information} have applied machine learning methods to EDGAR data and have provided useful results that can be used in real-world financial applications.

\section{Data}

Our overall dataset is comprised of 81 manually annotated \textit{10-K} reports, with a total of 1355 sentences holding 4522 entities and 3841 relations. A \textit{10-K} report is a comprehensive report filed annually by a publicly-traded company about its financial performance. The dataset was scraped from the EDGAR (Electronic Data Gathering, Analysis, and Retrieval system) database, which is maintained by the U.S. Securities and Exchange Commission and where all in the U.S. publicly listed companies have to submit a variety of financial reports.

As a first pre-processing step we tokenize the reports on a sentence level and then on a word level. Thereafter, we identify monetary values, the scale they are in (e.g. billion) and which currency unit is attached to it (usually U.S. Dollars) through rule-based string matching heuristics. This allows us to filter our reports for sentences containing such entities, as the focus of our dataset is to extract numeric monetary values and their corresponding key performance indicator. Given these sentences, we annotate word-level entities and relations. 

These annotations were generated by a team of four annotators, led by an auditing expert. The various entities and their annotations guideline, shown in Table \ref{tab:entities}, as well as the allowed relation matrix, which restricts which entities can be linked together, shown in Table \ref{tab:relations}, were defined in consultation with a wider team of auditors, who provided valuable feedback during the earlier stages of our project. After completing the annotations, the aforementioned senior auditing expert re-annotated 41 documents previously annotated by the other three annotators. 
The resulting inter-annotator agreement in terms of \textit{Cohen's Kappa}\footnote{The range of Cohen's Kappa lies between 1 (complete agreement of annotators) and -1 (complete disagreement of annotators), with 0 being no agreement between annotators or only what is to be expected by chance. See \cite{sim2005kappa} for an in-depth explanation on how to interpret \textit{Cohen's Kappa}.} \cite{cohen1960coefficient} on the word level is 0.7037. We further discuss this value and further insights regarding this inter-annotator agreement in Section~\ref{section:experiments} and the accompanying Table~\ref{tab:kappa}, which documents various \textit{Cohen's Kappa} scores.

\begin{table}[t]
\scriptsize
\renewcommand\tabularxcolumn[1]{m{#1}}
\rowcolors{2}{gray!10}{white}
\begin{tabularx}{\linewidth}{l@{\hspace{1em}}r@{\hspace{1em}}X} % p{11.5cm} p{12cm}
%\scriptsize
\toprule
\rowcolor{white} Entity &  Support & Description / Annotation Guideline \\
\midrule
kpi             & $1341$ & Key Performance Indicators expressible in numerical and monetary value, e.g. revenue or net sales.\\
cy              & $1211$ & Current Year monetary value of a KPI
% , almost always paired with some kind of currency token, e.g. \euro{} or EUR, and often a token signifying that the value is in thousands, millions, or even billions
.
\\
py              & $619$ & Prior Year monetary value of a KPI.\\
py1             & $307$ & 2 Year Past Value of a KPI \\
increase        & $35$ & Increase of a KPI from the previous year to the current year.\\
increase-py     & $15$ & Analogous to increase, but from py1 to py.\\
decrease        & $23$ & Decrease of a KPI from the previous year to the current year.\\
decrease-py     & $11$ & Analogous to decrease, but from py1 to py.\\
thereof           & $507$ & Represents a \textit{subordinate} KPI, i.e. if a KPI is part of another, broader KPI.\\
% davon-cy        & $8443$ & Current Year value of a thereof KPI.\\
% davon-py        & $4382$ & Prior Year value of a thereof KPI.\\
% davon-increase  & Increase from the previous year to the current year of a thereof KPI.\\
% davon-decrease  & Decrease from the previous year to the current year of a thereof KPI.\\
attr            & $272$ & Attribute that further describes a KPI.\\
kpi-coref       & $11$ & A co-reference to a KPI mentioned in a previous sentence.\\
false-positive  & $170$ & Captures tokens that are similar to other entities, but are explicitly not one of them, e.g. when the writer of the report forecasts next year's revenue.\\
\bottomrule
\end{tabularx}

% [('kpi', 16849),
%  ('cy', 11498),
%  ('davon', 8827),
%  ('davon_cy', 8443),
%  ('py', 5057),
%  ('davon_py', 4382),
%  ('increase', 356),
%  ('decrease', 230)]
\caption{Description and support of all entity types in the complete dataset, excluding the \textit{none} type.}
\label{tab:entities}
\end{table}

We randomly split the pre-processed dataset on the document
level into a training, validation, and test set, encompassing
969, 146, and 240 sentences each. 

The KPI-EDGAR dataset and code to reproduce the results herein are available at GitHub\footnote{See \href{https://github.com/tobideusser/kpi-edgar}{github.com/tobideusser/kpi-edgar}}.

\section{Methodology}
\label{section:methodology}

In this section, we shortly review the key points of KPI-BERT, a model tailored for key performance indicator extraction, introduced and illustrated in much greater detail in \cite{kpi-bert}
Thereafter, one span-level approach is briefly touched upon, namely the SpERT model introduced by \cite{eberts2019span}. 
We then provide two further baselines building on EDGAR--W2V \cite{loukas2021edgar} and GloVe \cite{pennington2014glove}, leveraging a similar setup like \cite{kpi-bert}.
These four models will be the baselines we provide to other researchers to benchmark their model against on KPI-EDGAR.

Finally, we describe our proposed adjustments to the prevalent F$_1$ metric to better capture the actual performance of machine learning models on relation extraction from financial documents.

\subsection{KPI-BERT}
\label{subsection:kpibert}

The KPI-BERT \cite{kpi-bert} model has three main building blocks: A BERT-based Sentence Encoder, a Named Entity Recognition (NER) Decoder, and a Relation Extraction (RE) Decoder.

\begin{table}[t]
\scriptsize
\setlength\tabcolsep{6pt}
\begin{subfigure}[b]{\linewidth}
\centering
\rowcolors{2}{gray!10}{white}
\begin{tabular}{lccccccccc}
\toprule
{} &  kpi  &  cy &  py &  py1 & increase & increase-py \\
\midrule
kpi             & - & 1:1 & 1:1 & 1:1 & 1:1 & 1:1 \\
cy             & 1:1 & - & - & - & - & - \\
py             & 1:1 & - & - & - & - & - \\
py1            & 1:1 & - & - & - & - & - \\
increase       & 1:1 & - & - & - & - & - \\
increase-py    & 1:1 & - & - & - & - & - \\
decrease       & 1:1 & - & - & - & - & - \\
decrease-py    & 1:1 & - & - & - & - & - \\
thereof        & n:1 & 1:1 & 1:1 & 1:1 & 1:1 & 1:1 \\
attr           & n:1 & - & - & - & - & - \\
kpi-coref      & - & 1:1 & 1:1 & 1:1 & 1:1 & 1:1 \\
false-positive & - & - & - & - & - & - \\
\bottomrule
\end{tabular}
\caption{}
\end{subfigure}

\setlength\tabcolsep{2pt}
\begin{subfigure}[b]{\linewidth}
\centering
\rowcolors{2}{gray!10}{white}
\begin{tabular}{lccccccc}
\toprule
{} & decrease & decrease-py & thereof & attr & kpi-coref & false-positive \\
\midrule
kpi            & 1:1 & 1:1 & 1:n & 1:n & - & - \\
cy             & - & - & 1:1 & - & 1:1 & - \\
py             & - & - & 1:1 & - & 1:1 & - \\
py1            & - & - & 1:1 & - & 1:1 & - \\
increase       & - & - & 1:1 & - & 1:1 & - \\
increase-py    & - & - & 1:1 & - & 1:1 & - \\
decrease       & - & - & 1:1 & - & 1:1 & - \\
decrease-py    & - & - & 1:1 & - & 1:1 & - \\
thereof        & 1:1 & 1:1 & - & - & n:1 & - \\
attr           & - & - & - & - & n:1 & - \\
kpi-coref      & 1:1 & 1:1 & 1:n & 1:n & - & - \\
false-positive & - & - & - & - & - & - \\
\bottomrule
\end{tabular}
\caption{}
\end{subfigure}

% \rowcolors{2}{gray!10}{white}
% \centering
% \begin{tabular}{l@{\hspace{\tabcolsep}}c@{\hspace{\tabcolsep}}c@{\hspace{\tabcolsep}}c@{\hspace{\tabcolsep}}c@{\hspace{\tabcolsep}}c@{\hspace{\tabcolsep}}c@{\hspace{\tabcolsep}}c@{\hspace{\tabcolsep}}c}
% \toprule
% {} &  kpi &  cy &  py & py1 & & increase & increase-py & decrease & decrease-py & thereof & \\
% \midrule
% kpi            & - & 1:1 & 1:1 & 1:1 & 1:1 & 1:n & - & - \\
% cy             & 1:1 & - & - & - & - & - & - & - \\
% py             & 1:1 & - & - & - & - & - & - & - \\
% increase       & 1:1 & - & - & - & - & - & - & - \\
% decrease       & 1:1 & - & - & - & - & - & - & - \\
% davon          & n:1 & - & - & - & - & - & 1:1 & 1:1 \\
% davon-cy       & - & - & - & - & - & 1:1 & - & - \\
% davon-py       & - & - & - & - & - & 1:1 & - & - \\
% \bottomrule
% \end{tabular}
\caption{Comprehensive overview of all allowed relations and their uniqueness. ``1:1'': One entity of type 1 can only be linked to one entity of type 2, ``1:n'': One entity of type 1 can be linked to many entities of type 2. ``-'': No relation possible.}
\label{tab:relations}
\end{table}

Given a WordPiece \cite{schuster2012japanese} tokenized input sentence we leverage a pre-trained BERT \cite{devlin2018bert} model to obtain our encoded token embedding. Thereafter, we apply a pooling function, which creates word representations by combining their individual subword embeddings. Both the more basic pooling functions \texttt{max} and \texttt{mean} as well as the specifically by \cite{kpi-bert} introduced  trainable recurrent neural network (RNN) pooling mechanism building on a bidirectional gated recurrent unit (GRU) \cite{cho2014properties} are evaluated in this context.

The NER decoder module classifies named entities within the sentence building on the IOBES annotation scheme. IOBES tagging functions by prepending all entity classes with the prefixes \textit{I-} (inside), \textit{B-} (begin), \textit{E-} (end) and \textit{S-} (single), or \textit{O} (outside) for tokens not belonging to the entity. Therefore, the number of possible IOBES entity tags is
\begin{align}\lvert \mathcal{E}_{\text{IOBES}} \rvert = 4 \left( \lvert \mathcal{E} \rvert - 1 \right) + 1, \end{align} 
where $\lvert \mathcal{E} \rvert$ is the number of entities, including the \textit{none} entity, as described in Table~\ref{tab:entities}. The inherently sequential nature of this task leads us to leverage a GRU in combination with conditional label masking to sequentially predict IOBES tags considering the previous predictions. 

To briefly elaborate, an embedding matrix $\mat{W}_{\text{label}} \in \RR^{\lvert \mathcal{E}_{\text{IOBES}} \rvert \times u}$ holding learnable $u$-dimensional embeddings of all IOBES entity types is defined and is concatenated to pooled BERT word embeddings $(\mat{e}_1, \mat{e}_2, \dots, \mat{e}_m)$, yielding the decoding input representation of any word $j$,
\begin{align}
    \mat{z}_j = \left[ \mat{e}_j; \mat{w}_{j-1}^\text{label}\right],
\end{align}
where $\mat{w}_{j-1}^\text{label} \in \RR^u$ is the embedding of the formerly predicted IOBES tag and $\mat{w}_0^\text{label}$ is defined as the \textit{O} (outside) label embedding.

Afterwards, $\mat{z}_j$ and the previous hidden state $\mat{h}_{j-1}$ is passed to a GRU, resulting in
\begin{align}
    \mat{h}_j = \text{GRU}\left( \mat{z}_j, \mat{h}_{j-1}\right).
\end{align}
We linearly transform $\mat{h}_j$ combined with masking out impossible tag predictions, e.g., an \textit{I} (inside) token without a previous \textit{B} (begin) token, and applying softmax to arrive at the IOBES tag predictions for word $j$:
\begin{align}
    \mat{\hat{y}}_j = \text{softmax}\left( \text{mask}\left(\mat{W}_{\text{seq}} \mat{h}_j + \mat{b}_{\text{seq}}\right) \right).
\end{align}

Finally, the RE Decoder takes the binary decision of whether two entities match or not. The entity sampling for this process is constrained by only passing allowed pairs, as defined in Table \ref{tab:relations}, to the decoder. The inputs to this decoder are the embeddings, defined for each entity $s$ as
\begin{align}
    \mat{e}(s) := \left[\text{pool}(\mat{e}_j, \mat{e}_{j+1}, \dots, \mat{e}_{j+k-1});\mat{w}_{k}^\text{width}\right],
    \label{eq:span_representation}
\end{align}
where $\quad \mat{e}(s) \in \RR^{d+v}$ and $\mat{w}_{k}^\text{width}$ is a span size embedding, which is taken from an embedding matrix $\mat{W}_{\text{width}} \in \RR^{l \times v}$ holding fixed-size learned embeddings of dimensionality $v$.

\subsection{Span-Level Baseline}
\label{subsection:spert}

At the core of this model, we utilize the same architecture as shown in \cite{eberts2019span}, where they titled the model SpERT, which we will use for this approach as well. Passing a byte-pair encoded input sentence of length $n$ through BERT \cite{devlin2018bert}, we obtain $n+1$ token embeddings, $(\mat{c}, \mat{e}_1, \mat{e}_2, \dots, \mat{e}_n)$, where $\mat{c}$ represents the context embedding for the whole sentence. 

In contrast to the approaches in sections \ref{subsection:kpibert} and \ref{subsection:furtherbaselines}, we classify entire entity spans instead of individual tokens. Thus, we create all possible token subsequences from the encoded BERT output, up to a maximum span length $l$, set to $l=10$ to keep true to the original configuration of \cite{eberts2019span}. For example, given the input \texttt{(alea iacta est)} we form the spans \texttt{(alea)}, \texttt{(iacta)}, \texttt{(est)}, \texttt{(alea, iacta)}, \texttt{(iacta, est)}, \texttt{(alea, iacta, est)}. Thereafter, each span is classified into entity types, and each possible entity pair receives a prediction whether or not it is linked together. 

Different from \cite{eberts2019span}, we additionally filter overlapping entity spans by removing the ones with a lower classification score, as overlapping spans are simply not possible in our application of extracting KPI's from financial documents. To illustrate, take the example above and let us assume that the span \texttt{(alea)} and \texttt{(alea, iacta)} are both classified as an entity, the first with a score of 0.75 and the latter with 0.5. In this approach, we would only keep the entity \texttt{(alea)}.

\subsection{Further Token-Level Baselines}
\label{subsection:furtherbaselines}

As defined in the methodology section \ref{section:methodology}, the introduced joint NER and RE model consists of three building blocks: a sentence encoder, NER decoder, and RE decoder. For generating further baselines, the sentence encoder of the described model is replaced by EDGAR--W2V \cite{loukas2021edgar} and GloVe \cite{pennington2014glove}.
The rest of the architecture is left unchanged. For these baselines, the sentences are tokenized into words and not WordPiece tokens.

The word embeddings are extracted from the available pre-trained embeddings uploaded by their respective authors. For those words which are not present in the vocabulary, a random vector is used as a substitute. 

\subsection{Adjusted F$_1$ Metric}
\label{subsection:adjustedf1}

To better capture the actual performance of predictions during the relation extraction from financial documents, we introduce an adjusted F$_1$ metric that allows for relations to be partially correct.

Given the following example sentence and ground truth relations from a financial document,
\begin{tcolorbox}[notitle,boxrule=0pt,
boxsep=0pt,left=0.6em,right=0.6em,top=0.5em,bottom=0.5em,
colback=gray!10,
colframe=gray!10]
``In 2021 and 2020 the $\underset{\text{{\color{fhgreen}{kpi}}}}{\text{{\color{fhgreen}{total net revenue}}}}$ was \$$\underset{\text{{\color{fhgreen}{cy}}}}{\text{{\color{fhgreen}{100}}}}$ million and \$$\underset{\text{{\color{fhgreen}{py}}}}{\text{{\color{fhgreen}{80}}}}$ million, respectively.''

\hfill \break
$\underset{\text{{\color{fhgreen}{kpi}}}}{\text{{\color{fhgreen}{total net revenue}}}} - \underset{\text{{\color{fhgreen}{cy}}}}{\text{{\color{fhgreen}{100}}}}$,
$\underset{\text{{\color{fhgreen}{kpi}}}}{\text{{\color{fhgreen}{total net revenue}}}} - \underset{\text{{\color{fhgreen}{py}}}}{\text{{\color{fhgreen}{80}}}}$,
\end{tcolorbox}

\noindent we found that predictions often omit a part of non-numerical entities like the KPI entity. In this example, predicting \textit{net revenue} instead of \textit{total net revenue} leads to a strict, conventional F$_1$ score of zero for both relations, even though the essential information was correctly extracted.

First, we define $o_i$ as the overlap/intersection of an entity prediction $i$ and its ground truth:

\begin{align}
    o_i := \left| e_{i, \text{pred}} \cap e_{i, \text{gt}} \right| , \label{eq1}
\end{align}

\noindent where $e_{i, \text{pred}}$ and $e_{i, \text{gt}}$ is the set of all token identifiers for the entity prediction and ground truth, respectively, and the operation $|\cdot|$ calculates the size of a given set.

The \textit{true positives} ($\text{tp}$), \textit{false negatives} ($\text{fn}$), and \textit{false positives} ($\text{fp}$) of a relation $r$ between entity $i$ and $j$ are then calculated by:

\begin{align}
    \text{tp}_r &= \frac{1}{2} \left( \frac{o_i}{n_{i, \text{gt}}} + \frac{o_j}{n_{j, \text{gt}}} \right) \\
    \text{fn}_r &= 1 - \text{tp}_r \\
    \text{fp}_r &= \frac{1}{2} \left( \frac{n_{i, \text{pred}} - o_i}{n_{i, \text{pred}}} +  \frac{n_{j, \text{pred}} - o_j}{n_{j, \text{pred}}} \right), \label{eq4}
\end{align}

\noindent where $n_{i, \cdot} := |e_{i, \cdot}|$ .
\addtocounter{footnote}{+1}
With this, we can calculate precision, recall, and the F$_1$ score of the relation $r$ in the conventional way:\footnotetext{Please note that no extensive grid search and finetuning was performed in this context, as the aim of this work is to introduce a new dataset and a better suited metric and not to find the model best suited for this task. Accordingly, these results should be seen as baselines for further research.}

\begin{align}
    \text{precision}_r &= \frac{\text{tp}_r}{\text{tp}_r + \text{fp}_r} \\
    \text{recall}_r &= \frac{\text{tp}_r}{\text{tp}_r + \text{fn}_r} \\
    \text{F}_{1,r} &= 2 \cdot \frac{\text{precision}_r \cdot \text{recall}_r}{\text{precision}_r + \text{recall}_r}.
\end{align}

\addtocounter{footnote}{-1}
\begin{table}[t]
\centering
\scriptsize
\rowcolors{2}{gray!10}{white}
\begin{tabular}{lcc}
\toprule
\rowcolor{white} Model & Relation F$_1$ in \% & Adjusted Relation F$_1$ in \%\\
\midrule
KPI-BERT   & $\bm{22.68}$ & $\bm{43.76}$ \\

SpERT   & $20.95$ & $40.04$\\
EDGAR--W2V   & $6.13$ & $19.71$\\
GloVe   & $5.11$ & $17.18$\\

\bottomrule
\end{tabular}
\caption{
Model performances\protect\footnotemark, measured in the conventional as well as adjusted (see subsection \ref{section:methodology}) F$_1$ score, of the models described in sections \ref{subsection:kpibert}, \ref{subsection:furtherbaselines}, and \ref{subsection:spert} on our novel dataset KPI-EDGAR.
}
\label{tab:models}
\end{table}

\section{Experiments}
\label{section:experiments}

Herein, we briefly evaluate the unmodified KPI-BERT \cite{kpi-bert} (see subsection \ref{subsection:kpibert}), the same structure but with word embeddings from EDGAR--W2V \cite{loukas2021edgar} and GloVe \cite{pennington2014glove} 
instead of embeddings from BERT \cite{devlin2018bert} (see subsection \ref{subsection:furtherbaselines} for these two), and a competing span-based approach (see subsection \ref{subsection:spert}), as introduced by \cite{eberts2019span} and named SpERT, on our novel dataset. The results should be seen as a baseline on which further research can be benchmarked. We also look into a few prediction examples and study the strength and weaknesses of our proposed adjustment to the F$_1$ score.

Table~\ref{tab:models} shows the conventional as well as the adjusted F$_1$ score of KPI-BERT, SpERT, and baselines with EDGAR--W2V and GloVe as an encoder on KPI-EDGAR. A first and obvious observation is that approaches without a transformer as an encoder are severely lacking in performance compared to the ones utilizing such an encoder. Given the sequential and context based nature of detecting and extracting key performance indicators and their values, this result is what we believe can be expected from attention based models as their inherent ability is to model such relations. 

Looking at the adjusted F$_1$ score, it is, as expected, significantly higher than the conventional score, which begs the question: does it do a better job in actually measuring the success of the prediction?

\begin{table}[t]
\centering
\renewcommand\tabularxcolumn[1]{m{#1}}
\scriptsize
{\scriptsize
\begin{tabularx}{\linewidth}{lXr}
\toprule
& Sentence with predicted Entities & Relations \\
\midrule
(a) & Entity boundaries are debatable, leading to arguably correct predictions but a conventional F$_1$ score of zero & \\
\midrule
1 & (a) 
{\color{orange}[}{\color{blue}[$\underset{\text{kpi {\color{orange}kpi}}}{\text{Unrealized gains}}$]}
{\color{orange}totaled]}
\${\color{fhgreen}{[$\underset{\text{cy}}{\text{96}}$}]}
million in 2020, 
\${\color{fhgreen}{[$\underset{\text{cy}}{\text{88}}$}]}
million in 2019 and 
\${\color{fhgreen}{[$\underset{\text{cy}}{\text{73}}$}]}
million in 2018 [...].  & 
\tiny{\color{fhgreen}{\makecell[r]{
{\color{blue}kpi -- cy} \\
{\color{orange}kpi -- cy} \\
{\color{blue}kpi -- py} \\
{\color{orange}kpi -- py} \\
{\color{blue}kpi -- py1} \\
{\color{orange}kpi -- py1} 
}}} \\
\midrule
2 & As of December 31, 2020 and 2019, the Company’s
{\color{blue}[$\underset{\text{kpi}}{\text{Medicare Part D}}$}
{\color{orange}[$\underset{\text{kpi}}{\text{receivables}}$]}{\color{blue}]}
amounted to
\${\color{fhgreen}{[$\underset{\text{cy}}{\text{2.9}}$}]}
billion and 
\${\color{fhgreen}{[$\underset{\text{py}}{\text{2.3}}$}]}
billion , respectively . & 
\tiny{\color{fhgreen}{\makecell[r]{
{\color{blue}kpi -- cy} \\
{\color{orange}kpi -- cy} \\
{\color{blue}kpi -- py} \\
{\color{orange}kpi -- py}
}}} \\
\midrule
(b) & Difference in prediction and ground truth, however both are viable options. Nevertheless, the conventional and adjusted F$_1$ score will assign 0 here. & \\
\midrule
3 & As a result, we recognized 
\${\color{fhgreen}{[$\underset{\text{cy}}{\text{50}}$}]} million of 
{\color{blue}{[$\underset{\text{kpi}}{\text{costs}}$}]}
primarily related to 
{\color{orange}{[$\underset{\text{kpi}}{\text{employee termination expenses and losses}}$}]}
from closing certain stores impacting both segments  &
\tiny{\color{fhgreen}{\makecell[r]{
{\color{blue}kpi -- cy} \\
{\color{orange}kpi -- cy} 
}}}\\
\bottomrule
\end{tabularx}
}

\caption{Several example sentences from the test set of KPI-EDGAR with joint named entity recognition and relation extraction results. {\color{fhgreen}{Green}}, {\color{blue}{Blue}} and {\color{orange}{Orange}} represent ``true positive'', ``false positive'', and ``false negative'' entity and relation classifications, respectively. The predictions were generated by KPI-BERT \cite{kpi-bert}.
}
\label{tab:examples}
\end{table}

To answer this question, Table \ref{tab:examples} illustrates a few examples taken directly from the test set of KPI-EDGAR. Subtable (a) highlights sentences where our adjusted F$_1$ score enables the metric to reflect the actual performance of a model. These examples also reveal a critical issue annotators and thus, machine learning models have to face when extracting KPIs from financial documents: Where are the exact boundaries of an entity? What are relevant pieces of information that still belong to said entity and what is less important? Often, two expert auditors will have a differing opinion on where to put these boundaries. 
This inherently fuzzy and noisy process can also be observed when looking at Table~\ref{tab:kappa}, where various Cohen's Kappa scores are reported. As shown there, annotators agree strongly on where numeric entities like \textit{cy} or \textit{py} lie, but the location and especially borders of non-numeric entities like \textit{kpi} or \textit{thereof} are much more up to debate. 

\begin{table}[t]
\centering
\footnotesize
\rowcolors{2}{gray!10}{white}
\begin{tabular}{lc}
\toprule
\rowcolor{white}Type & Cohen's Kappa\\
\midrule
All words   & 0.7037 \\[3pt]

Only entities   & 0.4885 \\[3pt]

Entity: kpi   & 0.0822 \\
Entity: cy   & 0.5972 \\
Entity: py   & 0.6657 \\
Entity: py1   & 0.6095 \\
Entity: increase   & 0.7419 \\
Entity: increase-py   & 0.5556 \\
Entity: decrease   & 0.7713 \\
Entity: decrease-py   & -0.1538 \\
Entity: thereof   & 0.3053 \\
Entity: attr   & -0.1076 \\
Entity: kpi-coref   & -0.2592 \\
Entity: false-positive   & -0.7587 \\

\bottomrule
\end{tabular}
\caption{Cohen's Kappa \cite{cohen1960coefficient} scores of various input series. \textit{All words} includes all word tokens. \textit{Only entities} only includes word tokens that belong to an entity annotation. Types starting with \textit{Entity:} are calculated by only considering word tokens that are of such an entity.}
\label{tab:kappa}
\end{table}
Therefore, we believe that our proposed weighted F$_1$ will at least allow for some noise in the predicted boundaries and thus, be able to capture these fuzzy borders more accurately, especially when compared to the conventional strict F$_1$ score.

Nevertheless, the adjusted F$_1$ is unable to fairly capture all variations in annotation and prediction difference, as shown in Subtable (b) of Table \ref{tab:examples}. However, it is much harder to actually detect that both options are viable in these cases. It would require the metric to have an in-depth auditing knowledge that our proposed weighting scheme simply can not have.

\section{Conclusion and Future Work}

In this paper, we introduce KPI-EDGAR, a financial document dataset for joint named entity recognition and relation extraction, based on annual financial reports submitted to the EDGAR database by publicly-traded companies and manually annotated by us. Moreover, we present an adjusted F$_1$ score, based on a weighting scheme, that allows for relations to be partially correct in their entity boundaries.

We find that this metric can alleviate the problem of fuzzy boundaries of entities and thus relations during the extraction of key performance indicators from financial reports and is therefore superior to the conventional F$_1$ score for measuring the success of the relation extraction task in this domain. As these fuzzy boundaries occur frequently in KPI-EDGAR, we propose that other researchers use this adjusted F$_1$ score when measuring model performances on it.

This study is part of a larger endeavor and long-time research project to advance information extraction processes in the financial text domain, where both machine learning researchers and audit experts work together on finding novel solutions to practical issues encountered during the audit process. In its wake, we plan to further improve availability, quality, and quantity of datasets. 
As a logical next step, we intend to build a named entity recognition dataset based on available GAAP (general accepted accounting principle) entities from the EDGAR database. Given the massive amount of freely available financial reports on EDGAR, it is also conceivable to train a custom language model tailored to handle such financial documents, as shown with EDGAR--W2V \cite{loukas2021edgar} but with a state-of-the-art transformer model like BERT \cite{devlin2018bert}. Additionally, in a further push to improve performance tracking, we will investigate dynamic ways to weight the metric, e.g., by incorporating the importance of each word that is part of an entity and relation.

\balance
\bibliographystyle{IEEEtran} 
\bibliography{bibliography}

\end{document}